\documentclass[wcp]{jmlr}
\usepackage{natbib}
\usepackage{longtable}% for long tables

\usepackage{booktabs}
\usepackage{algorithm}
\usepackage{algorithmic}
\usepackage{bm}
\jmlrvolume{80}
\jmlryear{2017}
\jmlrworkshop{ACML 2017}

\title[Learning RBM with DC]{Learning RBM with a DC programming Approach}

  \author{\Name{Vidyadhar Upadhya} \MakeLowercase{\Email{vidyadhar@ee.iisc.ernet.in}}\\
  \addr Electrical Engineering Dept, Indian Institute of Science, Bangalore 
  \AND
  \Name{P. S. Sastry} \Email{sastry@ee.iisc.ernet.in}\\
  \addr Electrical Engineering. Dept, Indian Institute of Science, Bangalore
 }

\DeclareMathOperator{\argmax}{argmax}
\DeclareMathOperator{\argmin}{argmin}
\newcommand{\beq}{\begin{equation}}
\newcommand{\eeq}{\end{equation}}
\newcommand{\beqa}{\begin{eqnarray}}
\newcommand{\eeqa}{\end{eqnarray}}
\newcommand{\ben}{\begin{enumerate}}
\newcommand{\een}{\end{enumerate}}

\newcommand{\llk}{\mathcal{L}}
\newcommand{\Real}{\mathbb{R}}

\newcommand{\lb}{\left(}
\newcommand{\rb}{\right)}
\newcommand{\ls}{\left[}
\newcommand{\rs}{\right]}

\newcommand{\vecv}{\mathbf{v}}

\newcommand{\vecb}{\mathbf{b}}
\newcommand{\vecc}{\mathbf{c}}
\newcommand{\vecw}{\mathbf{w}}
\newcommand{\vech}{\mathbf{h}}
\newcommand{\vecg}{\mathbf{g}}

\newcommand{\vecmu}{\bm{\mu}}
\newcommand{\veclambda}{\bm{\lambda}}
\newcommand{\vecW}{\mathbf{W}}

\newcommand{\Ex}{\mathbb{E}}

\newcommand{\sv}{\tilde{\vecv}}

\newcommand{\st}{\tilde{\theta}}

\newcommand{\non}{\nonumber}

\def  \wrt{\textit{w.r.t. }}
\begin{document}

\maketitle

\begin{abstract}
%Exact learning of an RBM is computationally expensive due to the intractable nature of the log-likelihood gradient. Most algorithms use a gradient estimate obtained through MCMC sampling. 
By exploiting the property that the RBM log-likelihood function is the difference of convex functions,
we formulate a stochastic variant of the difference of convex functions (DC) programming to minimize the negative log-likelihood. 
Interestingly, the traditional contrastive divergence algorithm is a special case of
the above formulation and the hyperparameters of the two algorithms can be chosen such that the amount of computation per mini-batch is identical. 
We show that for a given computational budget the proposed algorithm 
almost always reaches a higher log-likelihood more rapidly, compared to the
standard contrastive divergence algorithm. Further, we modify this algorithm to use the centered gradients
and show that it is more efficient and effective compared to the standard centered gradient algorithm on benchmark datasets.

\end{abstract}

\section{Introduction}
The Restricted Boltzmann Machines (RBM) ~\citep{smolensky1986information,freund1994unsupervised,hinton2002training} are among the basic building blocks of 
several deep learning models including Deep Boltzmann Machine (DBM) ~\citep{salakhutdinov2009deep} and Deep Belief Networks (DBN) ~\citep{Hinton06}.   
Even though they are used mainly as generative models, RBMs  can be suitably modified to perform
classification tasks also. 

The model parameters in an RBM are learnt by maximizing the 
log-likelihood. However, the gradient (\textit{w.r.t.} the parameters of the model) of the log-likelihood 
 is intractable since it contains an expectation 
term \textit{w.r.t.} the model distribution. This expectation is 
computationally expensive: exponential in (minimum 
of) the number of visible/hidden units in the model. Therefore, this expectation is approximated by taking an average 
over the samples from the model distribution. 
The samples are obtained using Markov Chain Monte Carlo (MCMC) methods which 
can be implemented efficiently by exploiting the bipartite connectivity structure
of the RBM. The popular Contrastive Divergence (CD) algorithm is a stochastic gradient ascent on log-likelihood using the estimate of gradient obtained through MCMC procedure. However, the gradient estimate based on  MCMC methods may be poor when the underlying density is high dimensional and 
can make the simple stochastic gradient descent (SGD) based algorithms to even diverge in some cases ~\citep{fischer2010empirical}.

There are two approaches to alleviate these issues.
The first is to design an efficient MCMC method to get good representative samples from the model distribution and thus to get a reasonably accurate estimate of
the gradient ~\citep{desjardins2010adaptive,tieleman2009using}. However, sophisticated MCMC methods are computationally intensive, in general. The second 
approach is to use the noisy estimate based on simple MCMC but employing
sophisticated optimization strategies like second-order gradient descent ~\citep{Martens2010DeepLV}, natural gradient ~\citep{metricfree}, stochastic spectral descent (SSD)~\citep{carlson2015stochastic},etc.
These sophisticated optimization methods often result in additional computational costs.

In this paper, we follow the second approach and propose an efficient optimization procedure based on the so called difference of convex 
functions programming (or concave-convex procedure), by exploiting 
the fact that the RBM log-likelihood is the difference of two convex functions.  
Even for such a formulation we still need the intractable gradient which is approximated with the usual MCMC based sampling. We refer to the proposed algorithm as the {\it stochastic- difference of convex functions programming}
(S-DCP) since the optimization involves stochastic approximation over the existing difference of convex functions programming approach.

What is interesting is that the resulting learning algorithm that we derive turns out to be an interesting and simple modification of the 
standard Contrastive Divergence, CD, which is the most popular algorithm for training RBMs. As a matter of fact, the CD can be seen to 
be a special case of our proposed algorithm. Thus our algorithm can be viewed as a small modification of CD which is theoretically 
motivated by looking at the problem as that of optimizing difference of convex functions. Although small, this modification to CD turns out 
to be important because our algorithm exhibits a much superior rate of convergence as we show through extensive simulations on benchmark data 
sets. Due to the similarity of our method with CD, it is possible to choose hyperparameters of the two algorithms such that the amount of 
computation per mini-batch is identical. Hence, we show that, for a fixed computational budget, our algorithm reaches a higher log-likelihood 
more rapidly compared to CD. 
Our empirical results provide a strong justification for preferring this method over the traditional SGD approaches.

Further, we modify the S-DCP algorithm to use the centered gradients (CG) as in  ~\cite{JMLR:v17:14-237}, motivated by
the principle that by removing the mean of the training data and the mean of the hidden activations from the 
visible and the hidden variables respectively the conditioning of the underlying optimizing problem can be improved ~\citep{Montavon2012}.
The simulation results on benchmark data sets indicate that models
learnt by S-DCP algorithm with centered gradients achieve better log-likelihood compared to the other standard methods.

It is important to note that the proposed method is a minorization-maximization algorithm and can be viewed as an instance of expectation maximization (EM) method. Since
RBM is a graphical model with latent variables,  many algorithms based on ML estimation, including CD, can be cast as a (generalized) expectation maximization (EM) algorithm. In itself, this EM view does not give any extra insight.

In fact, learning RBM using
EM, alternate minimization and maximum likelihood approach are all similar ~\citep{van1994boltzmann,amari1992information}. Therefore, the proposed
algorithm can be interpreted as a tool which provides better optimization dynamics to learn RBM with any of these approaches.

The rest of the paper is organized as follows. In section \ref{sec:background}, 
we first briefly describe the RBM model and the maximum likelihood (ML) learning approach for RBM. 
We explain the proposed algorithm, the S-DCP, in section \ref{sec:DCA}.
In section \ref{sec:experiments}, we describe the 
simulation setting and then present the
results of our study. Finally, we conclude the paper in 
section~\ref{sec:conclusions}.
\section{Background}
\label{sec:background}
\subsection{Restricted Boltzmann Machines}
The Restricted Boltzmann Machine (RBM) is an energy based model
with a two layer architecture, in which $m$ visible stochastic units $(\vecv)$ in one layer are connected to $n$ hidden stochastic units $(\vech)$ in the other layer ~\citep{smolensky1986information,freund1994unsupervised,hinton2002training}.
Connections within a layer (i.e., visible to visible and hidden to hidden) are absent and the connections between the layers are undirected.
This architecture is normally used as a generative model.
The units in an RBM can take discrete or continuous values. 
In this paper, we consider the binary case , i.e., $\vecv\in\{0,1\}^m$ and $\vech\in\{0,1\}^n$. The probability distribution represented by the model with parameters, $\theta$, is  
\beq
 p(\vecv,\vech\vert\theta)=e^{-E(\vecv,\vech;\theta)}/Z(\theta)\label{llk_eq}
 \eeq
 where, $Z(\theta)=\sum_{\vecv,\vech}e^{-E(\vecv,\vech;\theta)}$ is the so called partition function and 
$E(\vecv,\vech;\theta)$ is the energy function defined by
 \beq
 E(\vecv,\vech;\theta)=-\sum_{i,j}w_{ij} h_i\, v_j-\sum_{j=1}^{m} b_j\,v_j-\sum_{i=1}^{n} c_i\, h_i\non\\
 \eeq
  where $\theta=\{\vecw\in\Real^{n\times m},\vecb\in\Real^{m},\vecc\in\Real^{n}\}$ is the set of model parameters. The $w_{ij}$, the $(i,j)^{\text{th}}$ element of $\vecw$, is the weight of the connection between the $i^{\text{th}}$ hidden unit and the $j^{\text{th}}$ visible unit. The bias for the $i^{\text{th}}$ hidden unit and the $j^{\text{th}}$ visible unit are denoted as $c_i$ and $b_j$, respectively.  
\subsection{Maximum Likelihood Learning}\label{sec_ML}
The RBM parameters, $\theta$, are learnt through the maximization of the 
log-likelihood over the training samples. The log-likelihood, given one training sample ($\vecv$), is given by,
\beqa
\llk (\theta\vert \vecv)&{=}&\log\,p(\vecv\vert\theta)\non\\
&{=}&\log\,\sum_\vech p(\vecv,\vech\vert\theta)\non\\
&{=}&\log\,\sum_\vech e^{-E(\vecv,\vech:\theta)}- \log\,Z(\theta) \non \\
& \triangleq & ( g(\theta,\vecv) - f(\theta)) \label{ll_base}
\eeqa
where we define 
\beqa
g(\theta,\vecv)&=& \log\,\sum_\vech e^{-E(\vecv,\vech:\theta)}\non\\
f(\theta)&=& \log\,Z(\theta)=\log \,\sum_{\vecv',\vech}e^{-E(\vecv',\vech;\theta)}\label{f_g_def}
\eeqa
The optimal RBM parameters are to be found by solving the following optimization problem.
\beq
\theta^*=\argmax_\theta \llk (\theta\vert \vecv)= \argmax_\theta \,\,( g(\theta,\vecv)- f(\theta))\label{ll_opt}
\eeq
The stochastic gradient descent iteratively updates the parameters as,
\beq
\theta^{t+1}=\theta^{t}+\left.\eta\,\,\nabla_\theta\llk (\theta \vert \vecv)\right\vert_{\theta=\theta^t}\non
\eeq
One can show that ~\citep{hinton2002training,fischer2012introduction},
\beqa
\nabla_\theta\,g(\theta,\vecv)&=&-\frac{\sum_\vech e^{-E(\vecv,\vech:\theta)} \nabla_\theta\, E(\vecv,\vech;\theta)}{\sum_\vech e^{-E(\vecv,\vech:\theta)}} =-\Ex_{p(\vech\vert\vecv;\theta)}\ls \nabla_\theta\,E(\vecv,\vech;\theta)\rs\non\\
\nabla_\theta \,f(\theta)&=& -\frac{\sum_{\vecv',\vech} e^{-E(\vecv',\vech;\theta)}\nabla_\theta\, E(\vecv',\vech;\theta)}{\sum_{\vecv',\vech} e^{-E(\vecv',\vech;\theta)}}=-\Ex_{p(\vecv',\vech;\theta)}\ls\nabla_\theta\, E(\vecv',\vech;\theta)\rs\label{loglik_grad}
\eeqa
where $ \Ex_q$ denotes the expectation \wrt the distribution $q$. 
The expectation under the conditional distribution, $p(\vech\vert\vecv;\theta)$, for a given $\vecv$, has a closed form expression and hence, $\nabla_\theta\,g$ is easily evaluated analytically. However, expectation under 
the joint density,  $p(\vecv,\vech;\theta)$, is computationally intractable since the number of terms in the expectation summation grows exponentially with the 
(minimum of) the number of 
hidden units/visible units present in the model. Hence, sampling methods are used to obtain $\nabla_\theta\,f$. 
\subsection{Contrastive Divergence}\label{sec_CD_MCMC}
The 
contrastive divergence ~\citep{hinton2002training}, a popular algorithm to learn RBM, is based on Hastings-Metropolis-Gibbs sampling. In this 
algorithm, a single sample, obtained after running a Markov chain for $K$ steps, is used to approximate the expectation as,
\beqa
\nabla_\theta \,f(\theta)&=&-\Ex_{p(\vecv,\vech;\theta)}\ls\nabla_\theta\, E(\vecv,\vech;\theta)\rs\non\\
&=&-\Ex_{p(\vecv;\theta)}\Ex_{p(\vech\vert\vecv;\theta)}\ls\nabla_\theta\, E(\vecv,\vech;\theta)\rs\non\\
&\approx&-\Ex_{p(\vech\vert\sv^{(K)};\theta)}\ls\nabla_\theta\, E(\sv^{(K)},\vech;\theta)\rs\non\\
&\triangleq& \hat{f'}(\theta,\sv^{(K)})
\eeqa
Here $\sv^{(K)}$ is the sample obtained after $K$ transitions of the Markov chain (defined by the current parameter values $\theta$) initialized with the training sample $\vecv$. 
A detailed description of the method is given as Algorithm \ref{CDk}.
In practice, the mini-batch version of this algorithm is used.
\begin{algorithm}[tb]
   \caption{CD-$K$ update for a single training sample $\vecv$}\label{CDk}
\begin{algorithmic}
   \STATE {\bfseries Input:} $\vecv,K,\theta^{(t)},\eta$
   \STATE $\sv^{(0)}=\vecv$
   \FOR{$k=0$ {\bfseries to}  $K-1$}
   \STATE sample $h_i^{(k)}\sim p(h_i\vert\sv^{(k)},\theta), \forall i$
   \STATE sample $\tilde{v}_j^{(k+1)}\sim p(v_j\vert\vech^{(k)},\theta), \forall j$
   \ENDFOR
\STATE {\bfseries Output:} $\theta^{(t+1)}=\theta^{(t)}-\eta  \ls\hat{f'}(\theta,\sv^{(K)})-\nabla g(\theta,\vecv)\rs$      
\end{algorithmic}
\end{algorithm}
There exist many variations of this CD algorithm in the literature, such as 
 persistent (PCD)~\citep{tieleman2008training}, fast persistent (FPCD)~\citep{tieleman2009using}, population (pop-CD)~\citep{KrauseFI15}, and average contrastive divergence (ACD)~\citep{e18010035}. 
Another popular algorithm, parallel tempering (PT)~\citep{desjardins2010adaptive}, is also based on MCMC. 
\section{DC Programming Approach}\label{sec:DCA}
The RBM log-likelihood is the difference between the functions $f$ and $g$, both of which are log sum exponential
functions and hence
are convex. We can exploit this property to solve the optimization problem in \eqref{ll_opt}
more efficiently by using the method known as 
 the concave-convex procedure (CCCP) ~\citep{yuille2002concave} 
or the difference of convex functions programming (DCP) ~\citep{An2005}.

The DCP is an algorithm to solve optimization problems of the form,
\beq
\theta^{*}=\argmin_\theta F(\theta)=\argmin\limits_\theta \lb f(\theta)-g(\theta)\rb\label{CCP_DCA_base}
\eeq
where, both the functions $f$ and $g$ are convex and $F$ is smooth but non-convex. 
The DCP algorithm is an iterative procedure defined by
\beq
\theta^{(t+1)}=\argmin_\theta \lb f(\theta)-\theta^T \nabla g(\theta^{(t)})\rb\label{CCP_DCA_convx}
\eeq
Note that each iteration as given above, solves a convex optimization problem because the objective function on the RHS of (\ref{CCP_DCA_convx}) is convex in $\theta$. By differentiating this convex function and equating it to zero, we see that the iterates satisfy $\nabla f(\theta^{t+1})=\nabla g(\theta^t)$. This gives an interesting geometric insight into this procedure. Given the current $\theta^t$, if we can directly solve 
$\nabla f(\theta^{t+1})=\nabla g(\theta^t)$ for $\theta^{t+1}$, then we can use it. Otherwise we 
solve the convex optimization problem (as specified in eq. \eqref{CCP_DCA_convx}) at each iteration through some numerical procedure. 

In the RBM setting, $F$ corresponds to the negative log-likelihood function and the functions $f,g$ are as defined in \eqref{f_g_def}. In our method, we propose to 
solve the convex optimization problem given by 
\eqref{CCP_DCA_convx} by using gradient descent on $f(\theta)-\theta^T \nabla g(\theta^{(t)},\vecv)$. 
For this, we still need $\nabla f$ which is computationally intractable. We propose to use the sample based estimate (as in Contrastive Divergence) 
for this and do a fixed number (denoted as $d$) of SGD iterations to minimize $f(\theta)-\theta^T \nabla g(\theta^{(t)},\vecv)$. A detailed description of this proposed algorithm is given as  
Algorithm \ref{S-DCP}. In practice, the mini-batch version of this algorithm is used, which is given as Algorithm \ref{SDCP_minibatch}. We refer
to this algorithm as the {\textbf{\it{stochastic-DCP}}}~(S-DCP) algorithm.

\begin{algorithm}[tb]
   \caption{S-DCP update for a single training sample $\vecv$}\label{S-DCP}
\begin{algorithmic}
   \STATE {\bfseries Input:} $\vecv,\theta^{(t)},\eta,d, K'$
   \STATE Initialize $\st^{(0)}=\theta^{(t)},\sv^{(0)}=\vecv$
       \FOR{$l=0$ {\bfseries to}  $d-1$}
           \FOR{$k=0$ {\bfseries to}  $K'-1$}
               \STATE sample $h_i^{(k)}\sim p(h_i\vert\sv^{(k)},\st^{(l)}), \forall i$
	       \STATE sample $\tilde{v}_j^{(k+1)}\sim p(v_j\vert\vech^{(k)},\st^{(l)}), \forall j$
           \ENDFOR
           \STATE $\st^{(l+1)}=\st^{(l)}-\eta \ls\hat{f'}(\st^{(l)},\sv^{(K')})-\nabla g (\theta^{(t)},\vecv)\rs $  
           \STATE $\sv^{(0)}=\sv^{(K')}$\label{v0vk}
   \ENDFOR
\STATE {\bfseries Output:}  $\theta^{(t+1)}=\st^{(d)}$
   \end{algorithmic}
\end{algorithm}
By comparing Algorithm \ref{S-DCP} with Algorithm \ref{CDk} it is easily seen that S-DCP and CD are very similar when viewed as iterative 
procedures. As 
a matter of fact if we choose the hyperparameter $d$ in S-DCP as 1, then they are identical. Therefore CD 
algorithm turns out to be a special case of S-DCP.
Further, there is a similarity between the initialization steps of S-DCP and PCD. 
Specifically, the step $\sv^{(0)}=\sv^{(K')}$ in Algorithm \ref{S-DCP}, 
retains the last state of the chain, to use it as the initial state in the next iteration. 
It is important to note that
for each gradient descent step, unlike in PCD, in S-DCP we start the Gibbs chain on a given example. In other words, for each gradient descent
step on negative log-likelihood, we have an inner loop that is several gradient descent steps on an auxiliary convex function. 
It is only during this inner loop that a single Gibbs chain is maintained.

The hyperparameter $d$ in S-DCP, controls the number of descent steps in the inner loop 
that optimizes the convex function $f(\theta)-\theta^T \nabla g(\theta^{(t)},\vecv)$. Since we are using 
the SGD on a convex function it is likely to be better behaved and thus, S-DCP may find better descent directions 
on the log likelihood of RBM as compared to CD. This may be the case even when the gradient $\nabla f$ obtained through 
MCMC is noisy. This also means we may be able to trade more steps on this gradient descent with fewer number of iterations of the MCMC. This allows us more flexibility
in using a fixed computational budget. 

We further modify the proposed S-DCP to use the centered gradients. We follow the Algorithm 1 given in ~\cite{JMLR:v17:14-237}. The detailed description of the CS-DCP algorithm is given as Algorithm \ref{CSDP_alg}.

In standard DCP or CCCP, one assumes that the optimization problem on the RHS of (\ref{CCP_DCA_convx}) is 
solved exactly to show that the algorithm given by (\ref{CCP_DCA_convx}) is a proper descent procedure for 
the problem defined by (\ref{CCP_DCA_base}). However, all that we need to show this is to ensure that we get descent 
on $f(\theta)-\theta^T \nabla g(\theta^{(t)})$ in each iteration. In our case, each component of the gradient of $f$ and $g$ 
is an expectation of a binary random variable (as can be seen from eq. \eqref{loglik_grad}) and hence is bounded by one. Thus, euclidean norms of 
gradients of both $f$ and $g$ are bounded by $\sqrt{(mn + m + n)}$, which is the square root of the dimension of the parameter vector, $\theta$. Since both $f$ and $g$ 
are convex, this implies that gradients of $f$ and $g$ are globally Lipschitz (chapter 3,~\cite{Bubeck:2015}) with known Lipschitz constant. Hence, we can have a constant step-size 
gradient descent on 
$f(\theta)-\theta^T \nabla g(\theta^{(t)})$ that ensures descent on each iteration.

Since we are using a noisy estimate for the gradient in the inner loop as explained above, the standard convergence proof for CCCP is not really applicable for the S-DCP. At present we do not have a full convergence proof for the algorithm because it is not easy to obtain
good bounds on the error in estimating  $\nabla f$ through MCMC. However, the simulation results that we present later show that S-DCP is effective and efficient.

\begin{algorithm}[tb]
   \caption{S-DCP update for a mini-batch of size $N_B$}\label{SDCP_minibatch}
\begin{algorithmic}
   \STATE {\bfseries Input:} $V=[\vecv^{(0)},\vecv^{(1)},\ldots,\vecv^{(N_B-1)}],\theta^{(t)},\eta,d,K'$
   \STATE Initialize $\st^{(0)}=\theta^{(t)},V_T=V,\Delta\theta=0$
       \FOR{$l=0$ {\bfseries to}  $d-1$}
        \FOR{$i=0$ { \bfseries to} $N_B-1$}
\STATE $\sv^{(0)}=V_T[:,i]\quad\rightarrow[i^{\text{th}}\text{ column of } V_T]$
           \FOR{$k=0$ {\bfseries to}  $K'-1$}
               \STATE sample $h_i^{(k)}\sim p(h_i\vert\sv^{(k)},\st^{(l)}), \forall i$
	       \STATE sample $\tilde{v}_j^{(k+1)}\sim p(v_j\vert\vech^{(k)},\st^{(l)}), \forall j$
           \ENDFOR
           \STATE $ \Delta \theta=\Delta \theta+ \ls\hat{f'}(\st^{(l)},\sv^{(K')})-\nabla g (\theta^{(t)},\vecv^{(i)})\rs$
           \STATE $V_T[:,i]=\sv^{(K')}$
            \ENDFOR
           \STATE $\st^{(l+1)}=\st^{(l)}-\eta \frac{\Delta \theta}{N_B} $  
   \ENDFOR
\STATE {\bfseries Output:}  $\theta^{(t+1)}=\st^{(d)}$
   \end{algorithmic}
\end{algorithm}

\begin{algorithm}[tb]
   \caption{CS-DCP update for a mini-batch of size $N_B$}\label{CSDP_alg}
\begin{algorithmic}
   \STATE {\bfseries Input:} $V=[\vecv^{(0)},\vecv^{(1)},\ldots,\vecv^{(N_B-1)}],\theta^{(t)},\eta,d,K',\vecmu,\veclambda,\nu_\mu,\nu_\lambda$
   \STATE Initialize $\st^{(0)}=\theta^{(t)},V_T=V,\Delta\theta=0,V_n=V_T$
   \STATE Calculate $H_p[t,i]=p(h_t=1\vert V_T[:,i])\,\,\forall t,i$ \hspace{0.5cm}/*           
               $p(h_t=1\vert\vecv)=\sigma\lb (\vecv-\vecmu)^T w_{*t}+c_t\rb$/*
   \STATE Calculate $\vecmu_{batch}=\text{Column\_mean}(V_T)$\\
                 \hspace{1.5cm}   $\veclambda_{batch}=\text{Column\_mean}(H_p)$
       \FOR{$l=0$ {\bfseries to}  $d-1$}
        \FOR{$i=0$ { \bfseries to} $N_B-1$}
\STATE $\sv^{(0)}=V_n[:,i]\quad\rightarrow[i^{\text{th}}\text{ column of } V_n]$
           \FOR{$k=0$ {\bfseries to}  $K'-1$}
               \STATE sample $h_i^{(k)}\sim p(h_i\vert\sv^{(k)},\st^{(l)}), \forall i\quad$\hspace{1cm}/*           
               $p(h_i=1\vert\vecv)=\sigma\lb (\vecv-\vecmu)^T w_{*j}+c_j\rb$/*
	       \STATE sample $\tilde{v}_j^{(k+1)}\sim p(v_j\vert\vech^{(k)},\st^{(l)}), \forall j$ \hspace{1.5cm}/*           
               $p(v_j=1\vert\vech)=\sigma\lb w_{i*}(\vech-\veclambda)+b_i\rb$/*
           \ENDFOR
           \STATE $V_n[:,i]=\sv^{(K')},\,\,H_n[t,i]=p(h_t=1\vert\sv^{(K')})\,\,\forall t$
            \ENDFOR            
            \STATE $\vecb=\vecb+\nu_\lambda \vecW^{(l)}(\veclambda_{batch}-\veclambda)$\hspace{2cm}/* Re-parameterization */
            \STATE $\vecc=\vecc+\nu_\mu \vecW^{(l)}(\vecmu_{batch}-\vecmu)$
            \STATE $\vecmu=(1-\nu_\mu)\vecmu+\nu_\mu\vecmu_{batch}$\hspace{2cm}/* moving average with sliding factor*/
            \STATE $\veclambda=(1-\nu_\lambda)\veclambda+\nu_\lambda\veclambda_{batch}$
            \IF{$l==0$}
           \STATE $ \nabla_w \vecg =\frac{(V_T-\vecmu) (H_p - \veclambda)^T}{N_B}$ /* Gradient of $\vecg$ \wrt $w$ is evaluated only once*/
            \ENDIF
            \STATE $\nabla \vecW=\nabla_w \vecg-\frac{ (V_n-\vecmu) (H_n - \veclambda)^T}{N_B}$
            \STATE $\nabla \vecb= \text{Column\_mean}(V_T)-\text{Column\_mean}(V_n)$
            \STATE $\nabla \vecc= \text{Column\_mean}(H_p)-\text{Column\_mean}(H_n)$     
            \STATE $\vecW^{(l+1)}=\vecW^{(l)}+\eta \nabla \vecW$
            \STATE $\vecb^{(l+1)}=\vecb^{(l)}+\eta \nabla \vecb$
            \STATE $\vecc^{(l+1)}=\vecc^{(l)}+\eta \nabla \vecc$
            \STATE $\st^{(l+1)}=\{\vecW^{(l+1)},\vecb^{(l+1)},\vecc^{(l+1)}\}$         
   \ENDFOR
\STATE {\bfseries Output:}  $\theta^{(t+1)}=\st^{(d)}$
   \end{algorithmic}
\end{algorithm}

\subsection{Computational Complexity}\label{subsec_CC}
Let us suppose that the computational cost of one Gibbs transition is $T$ and that of evaluating $g$ (and also $\hat{f'}$) is $L$. The
computational cost of the CD-$K$ algorithm for a mini-batch of size $N_B$ is $(N_B(KT+2L))$. The S-DCP algorithm with $K'$ MCMC steps and $d$ inner loop
iteration has cost
$(d\, N_B(K' T+L)+N_B L)$. For the S-DCP algorithm, $N_B L$ is not multiplied by $d$ because $\nabla g (\theta^{(t)},\vecv^{(i)})$ is evaluated 
only once for all the samples in the mini-batch. The computational cost of both CD and S-DCP algorithms can be made equal by choosing $dK'=K$,
if we neglect the term $(d-1)L$. This is acceptable since, $L$ is much smaller than $T$, and $d$ is also small (Otherwise we can choose $K'$ and $d$ to satisfy $KT=dK'T+(d-1)L$ to make the
computational cost of both the algorithms identical).
\section{Experiments and Discussions}\label{sec:experiments}
In this section, we give a detailed comparison between the S-DCP/CS-DCP and other standard algorithms like CD, PCD, centered gradient (CG)\citep{JMLR:v17:14-237} and stochastic spectral descent (SSD)\citep{carlson2015stochastic}.
In order to see the advantages of the S-DCP  over  
these CD based algorithms, we compare them by keeping 
the computational complexity same (for each  
mini-batch). Due to this computational architecture, the learning speed in terms of actual time is proportional to speed in terms of iterations.
We analyse the learning behavior by varying the hyperparameters, namely, the learning rate and the batch size.
We also provide simulation results showing the effect of hyperparameters, $d$ and $K'$ on S-DCP learning. Further, the sensitivity
to initialization is also discussed.

\subsection{The Experimental Set-up}\label{subsec:exp_setup}
We consider four benchmark datasets in our analysis namely Bars \& Stripes ~\citep{mackay2003information}, Shifting Bar ~\citep{JMLR:v17:14-237}, MNIST\footnote{statistically binarized as in ~\citep{salakhutdinov2008quantitative}} ~\citep{LeCun:1998} and Omniglot{\scriptsize{\footnote{https://github.com/yburda/iwae/tree/master/datasets}}} ~\citep{Lake1332}.
The Bars \& Stripes dataset consists of patterns of size $D\times D$ which are generated as follows.
First, all the pixels in each row are set to zero or one with equal probability. Second, the 
pattern is rotated by $90$ degrees with a probability of $0.5$. We have used $D=3$, for which we get
$14$ distinct patterns. The Shifting Bar dataset consists of patterns of size $N$ where a set of $B$ consecutive pixels with 
cyclic boundary conditions are set to one and the others are set to zero. We have used $N=9$ and $B=1$,
for which we get $9$ distinct patterns.
We refer to these as small datasets. The other two datasets, MNIST and Omniglot 
have data dimension of $784$ and
we refer to these as large datasets. All these datasets are used in the literature to benchmark the RBM learning.

For small datasets, we consider RBMs with $4$ hidden units and for large datasets, we consider RBMs with $500$
hidden units.   
The biases of hidden units are initialized to zero and the weights are initialized to samples drawn from a Gaussian distribution 
with mean zero and standard deviation $0.01$. The biases of visible units are initialized to
the inverse sigmoid of the training sample mean.

We use multiple trials, where each trial starts with a particular initial configuration for weights and biases. 
For small datasets, we use $25$ trials and for large datasets we use $10$ trials.
In order to have a fair comparison, we make sure that all the algorithms start with the same initial setting.    
We did not use any stopping criterion. Instead we learn the RBM for a fixed number of epochs. The training is 
performed for $50000$ epochs for the small datasets and $200$ epochs for the large datasets. 
The mini-batch learning procedure is used and the training dataset is shuffled after every epoch. However, for small datasets, i.e., Bars $\&$ Stripes and Shifting Bar, full batch
training procedure is used. We use batch size of $200$ for large data sets, unless otherwise stated .

We compare the performance of S-DCP and CS-DCP with CD, PCD and CD with centered gradient (CG). We also compare it with the SSD algorithm. 
We keep the computational complexity of S-DCP and CS-DCP same as that of CD based algorithms by  
choosing $K,d$ and $K'$ as prescribed in Section \ref{subsec_CC}. Since previous works
stressed on the necessity using large $K$ for CD to get a sensible generative model ~\citep{salakhutdinov2008quantitative,carlson2015stochastic}, 
we use $K=24$ (with $d=6,K'=4$ for DCP) for large datasets and $K=12$ (with $d=3,K'=4$ for DCP) for small datasets. 
In order to get an unbiased comparison, we did not use momentum and weight decay for any of the algorithms.

For comparison with the centered gradient method, we use the Algorithm 1 given in ~\cite{JMLR:v17:14-237} which corresponds to $dd_s^b$ in their notation.
However we use CD step size $K=24$ compared to single step CD used in Algorithm 1 given in ~\cite{JMLR:v17:14-237}. The hyperparameters $\nu_\mu$ and $\nu_\lambda$
are set to $0.01$. The initial value of $\mu$ is set to mean of the training data and $\veclambda$ is set to $\mathbf{0.5}$. The CS-DCP algorithm also uses the same 
hyperparameter settings.

The SSD algorithm~\citep{carlson2015stochastic} is a  bound optimization
algorithm which exploits the convexity of functions, $f$ and $g$. Since, the proposed
algorithm in this paper is also designed to exploit the convexity of $f$ and $g$, we compare our results with that of the SSD algorithm.
It is important to note that SSD has an extra cost of computing the singular value decomposition of 
a matrix of size  $m\times n$ which costs $\mathcal{O}(mn \min(m,n))$ for each gradient update. Here, $m$ and $n$ are
the number of hidden and visible units, respectively. 
We have observed that for the small data sets the SSD algorithm results in divergence of the log-likelihood. We also observed
similar behavior for large data sets when small batch size is used. Therefore performance of the SSD algorithm
is not shown for those cases. Since SSD is shown to work well when the batch size is of the order $1000$, we compare 
S-DCP and CS-DCP under this setting. We use exactly the same settings as used in  ~\cite{carlson2015stochastic} and report the performance on the    
statistically binarized MNIST dataset. For the S-DCP, we use $d=6$ and $K'=4$ to match the CD steps of $K=25$ used in ~\cite{carlson2015stochastic}
and the batch size $1000$.

The performance comparison is based on the log-likelihood achieved on the test set. We show the mean of the Average Test Log-Likelihood (denoted as ATLL) over all trials. 
For small RBMs, the ATLL, is 
evaluated exactly as,
\beq
ATLL=\sum_{i=1}^N \log \, p(\vecv_\text{test}^{(i)}\vert\theta)
\eeq
However for large RBMs, we estimate the ATLL with annealed importance sampling ~\citep{neal2001annealed}
with $100$ particles and $10000$ intermediate distributions according to a linear temperature scale between $0$ and $1$. 

\subsection{Performance Comparison}
In this section, we present a number of experimental results to illustrate
the performance of S-DCP and CS-DCP in comparison with the other methods. In most of the experiments, we observe that the ATLL achieved by CS-DCP is
greater than that by CD and other variants. Further, it achieves this 
relatively early in the learning stage.

\begin{figure}[t!]
    \centering
        \includegraphics[width=0.45\textwidth]{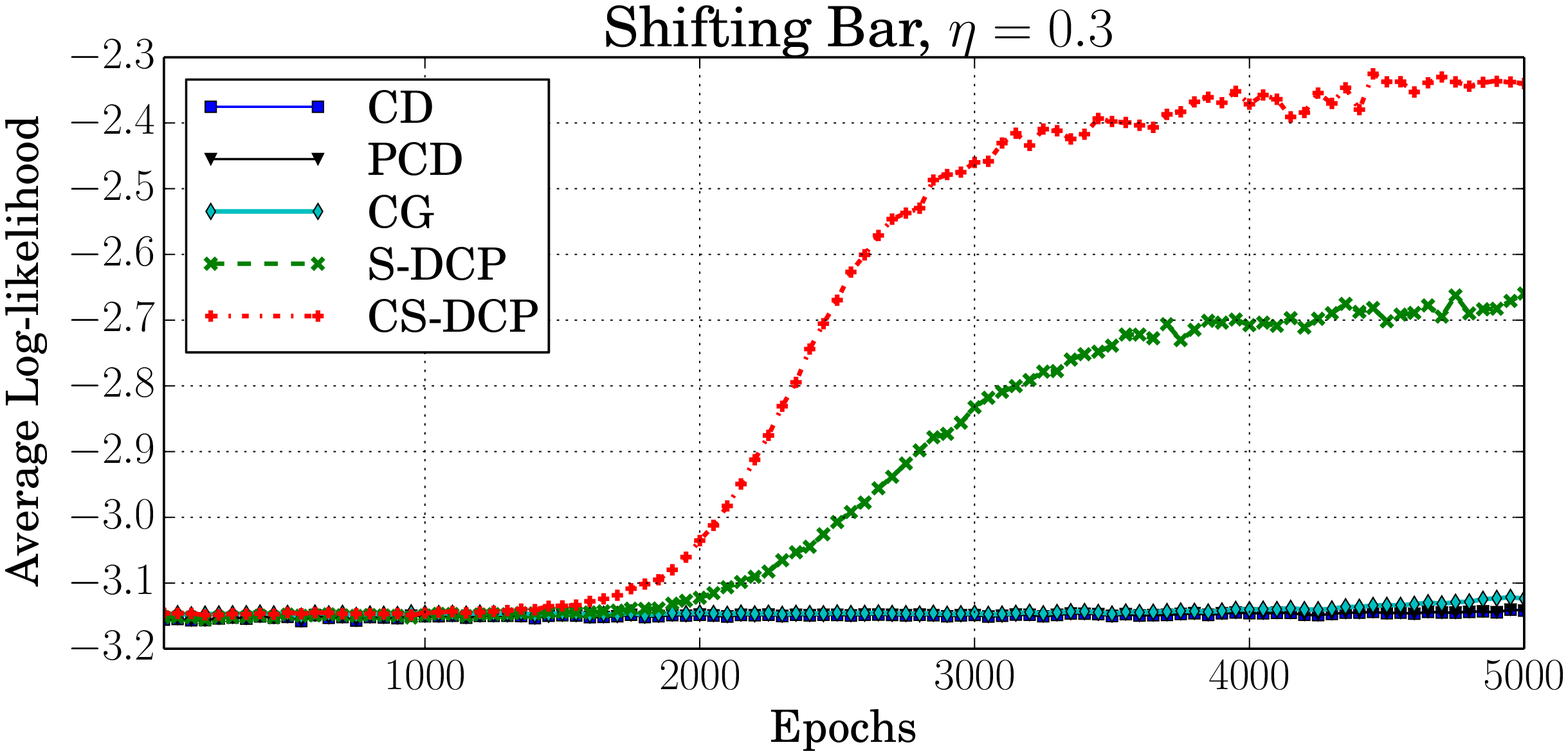}
        \includegraphics[width=0.45\textwidth]{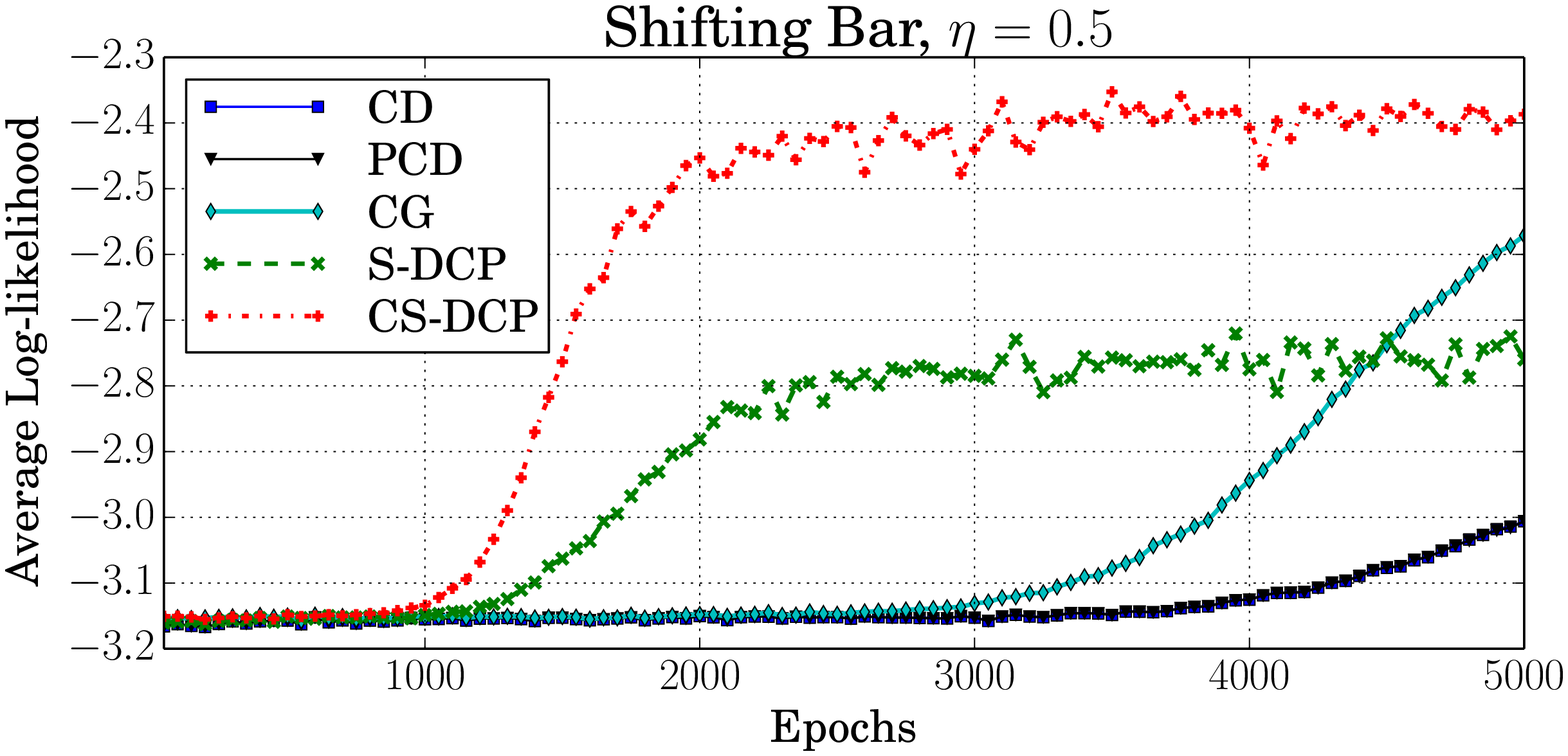}
    \caption{The performance of different algorithms on Shifting Bar dataset}\label{SDCPvsall_shiftbar}
\end{figure}
\begin{figure}[t!]
    \centering
        \includegraphics[width=0.47\textwidth]{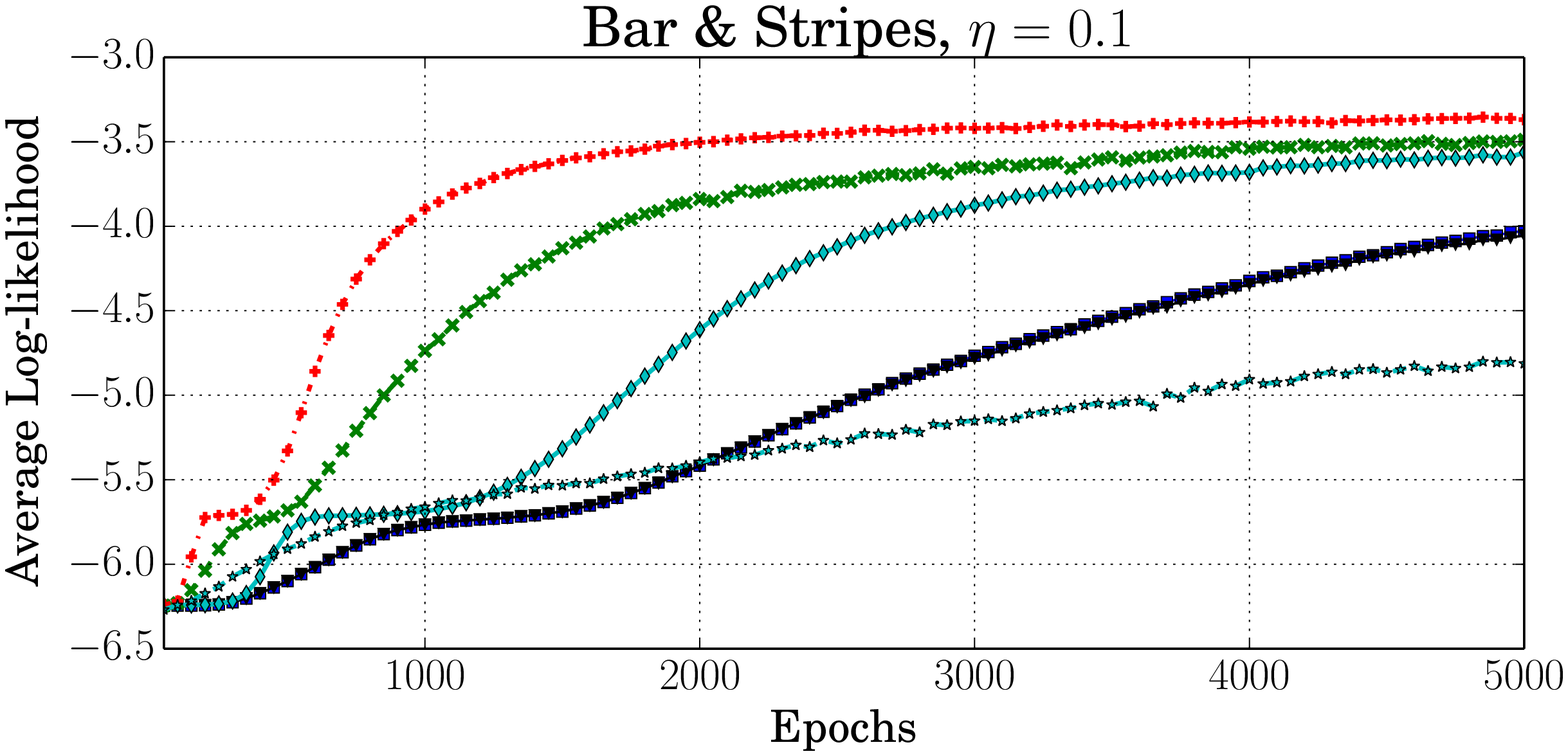}
        \includegraphics[width=0.47\textwidth]{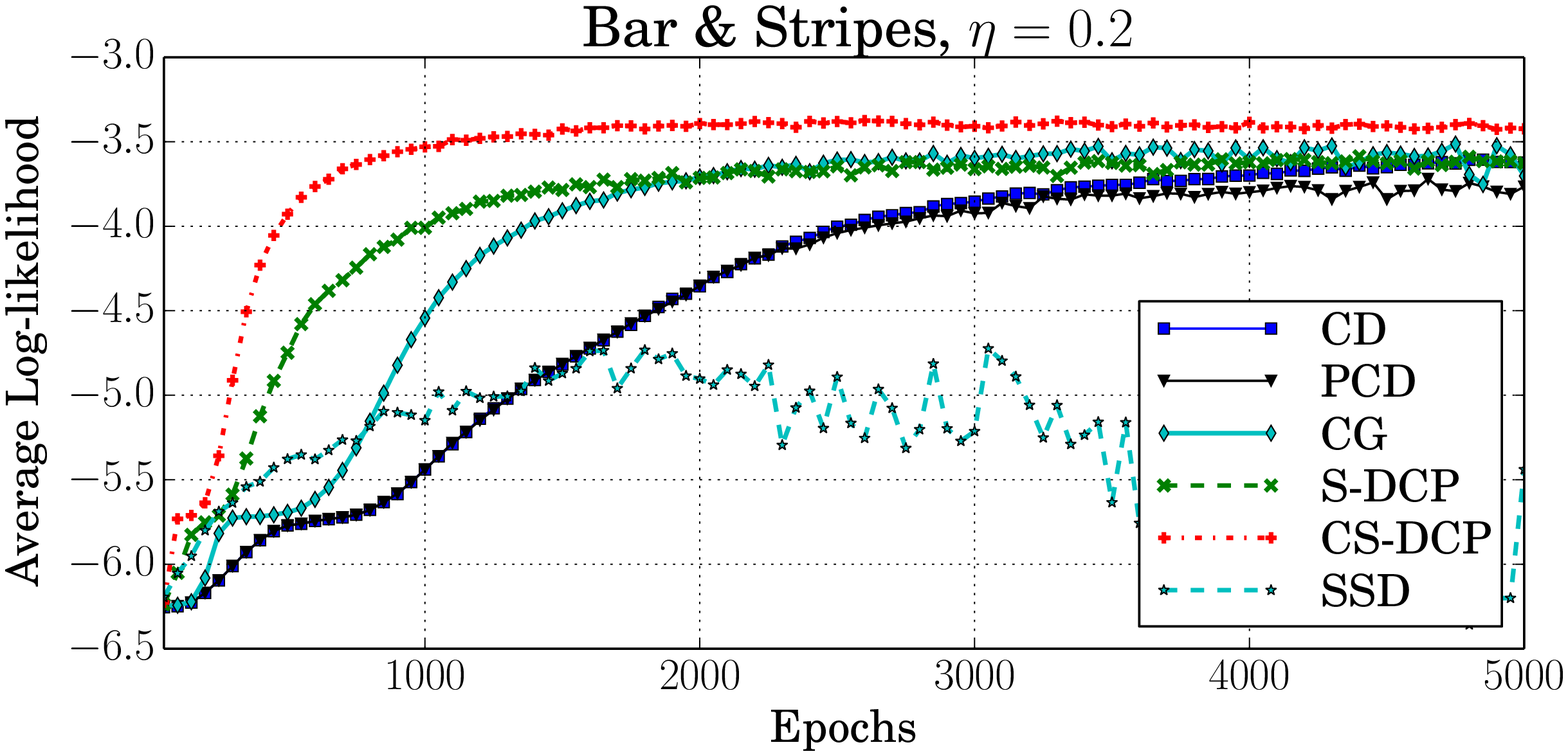}
    \caption{The performance of different algorithms on  Bars \& Stripes 
%     The vertical lines are tagged accordingly in the plot.
    }\label{SDCPvsall_barsstripes}
\end{figure}

\subsubsection{Small data sets}
Fig. \ref{SDCPvsall_shiftbar} and Fig. \ref{SDCPvsall_barsstripes} present results obtained with different algorithms on 
\textit{Bars \& Stripes} and \textit{Shifting Bar} data sets respectively\footnote{Please note that all the figures presented are best viewed in color.}. 
The theoretical upper bounds for the ATLL are $-2.6$ and $-2.2$ for  \textit{Bars $\&$ Stripes} and \textit{Shifting Bar} datasets respectively ~\citep{JMLR:v17:14-237}.

For the \textit{Shifting Bar} dataset we find the CD and PCD algorithms become stuck in a local minimum achieving ATLL around $-3.15$ for both the learning rates $\eta=0.3$ and $0.5$. We observe that the CG algorithm
is able to escape from this local minima when a higher learning rate of $\eta=0.5$ is used. However, S-DCP and CS-DCP algorithms are able to reach almost the 
same ATLL independent of the learning rate. We observed that in order to make the CD, PCD and CG algorithms learn 
a model achieving ATLL comparable to those of S-DCP or CS-DCP,  more than $30000$ gradient updates are required. This behavior is reported
in several other experiments~\citep{JMLR:v17:14-237}.

For the \textit{Bars \& Stripes} dataset, the CS-DCP/S-DCP and CG algorithms 
perform almost the same in terms of the achieved ATLL.
However CD and PCD algorithms are sensitive to the learning rate and fail to 
achieve the ATLL achieved by CS-DCP/S-DCP. Moreover, the speed of learning, 
indicated in the figure by the epoch at which the model achieve $90\%$ of the maximum ATLL,
shows the effectiveness of both CS-DCP and S-DCP algorithms. In specific, CG
algorithm requires around $2500$ epochs more of training compared to the
CS-DCP algorithm.

The experimental results on these two data sets indicate that both CS-DCP and S-DCP algorithms 
are able to provide better optimization dynamics compared to the other standard algorithms and they converge faster.

\subsubsection{Large data sets}
\begin{figure}[t!]
    \centering
        \includegraphics[width=0.44\textwidth]{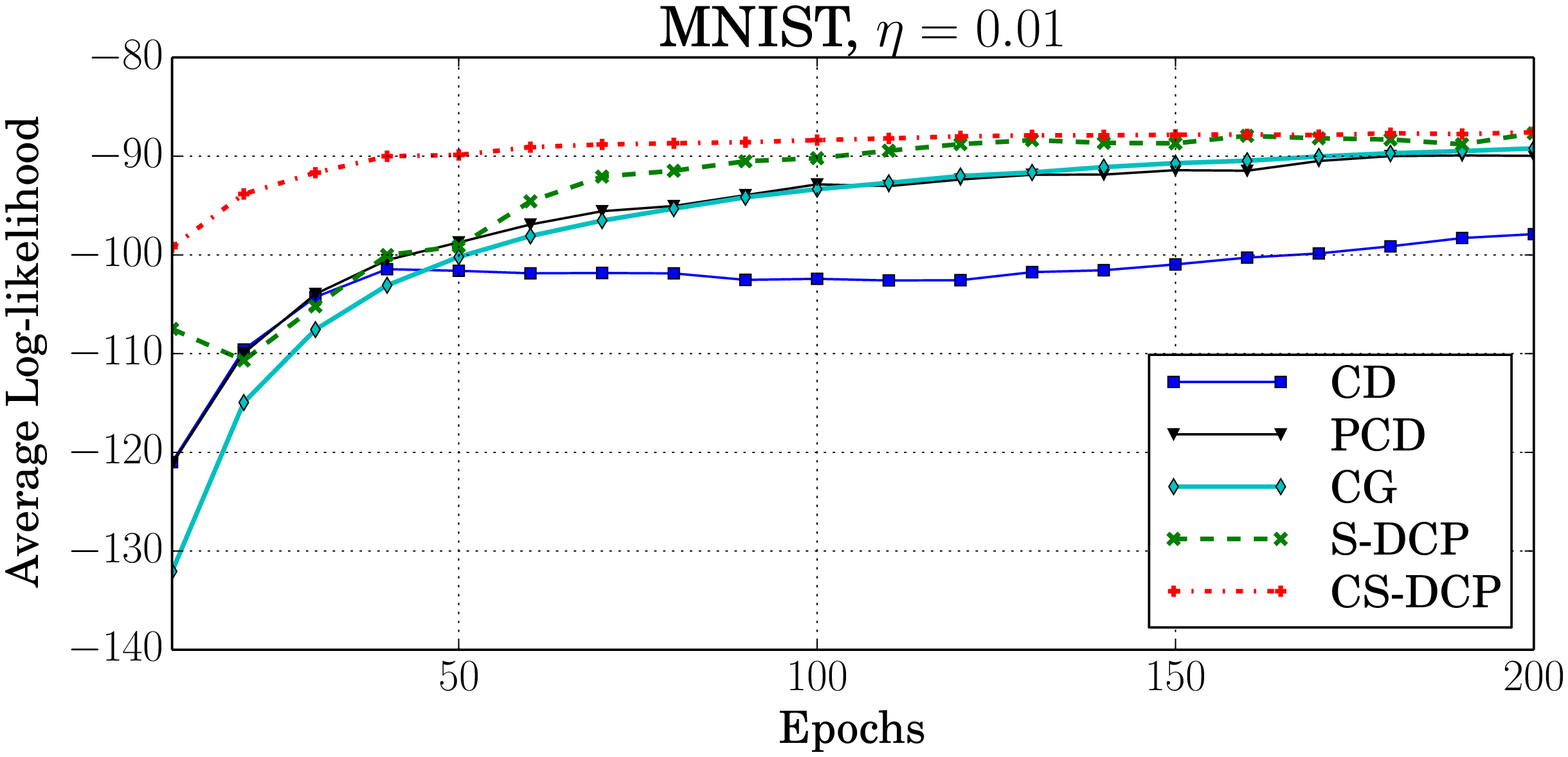}
        \includegraphics[width=0.44\textwidth]{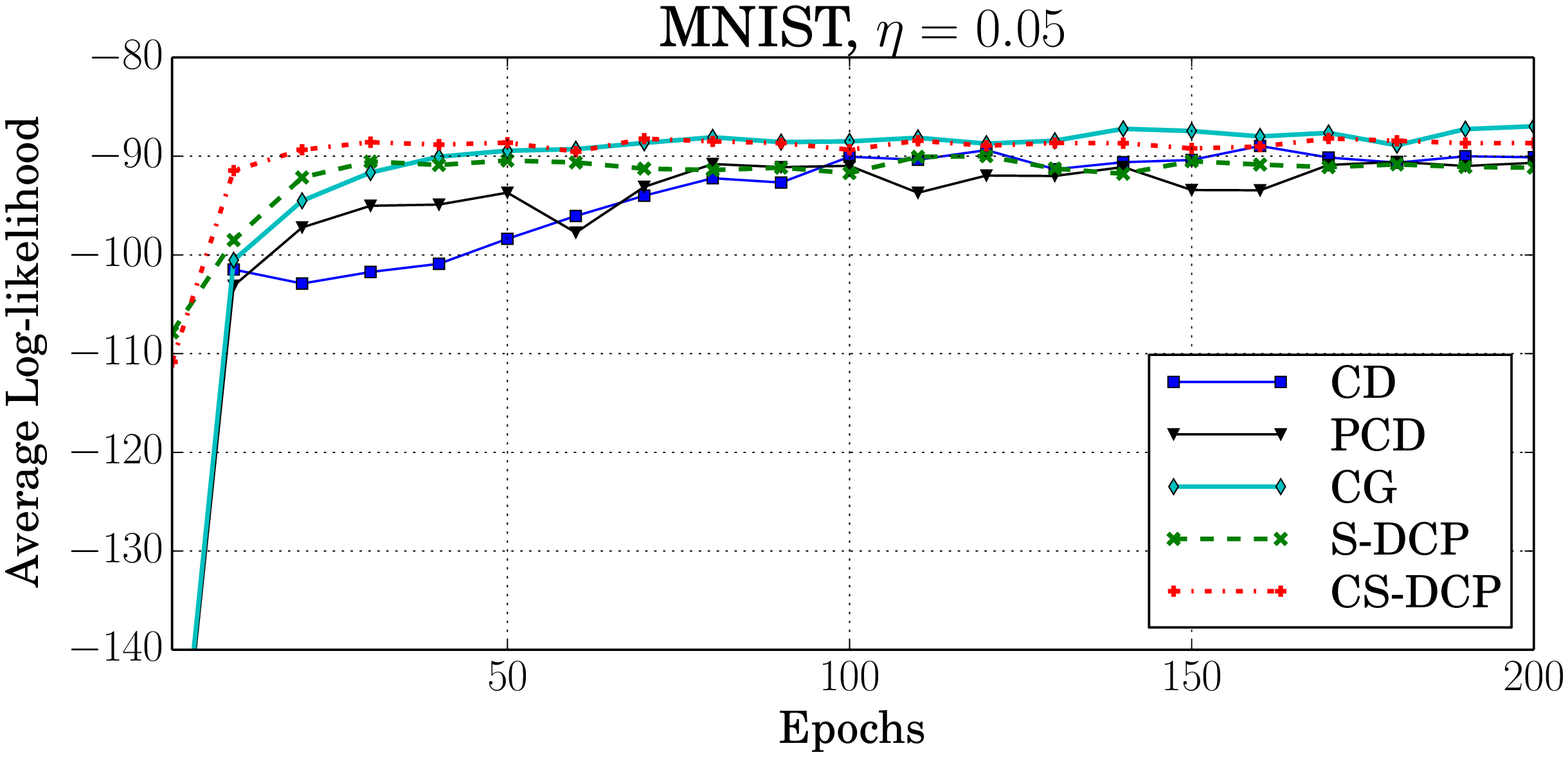}
    \caption{The performance of different algorithms on MNIST dataset 
%     The vertical lines are tagged accordingly in the plot
    .}\label{SDCPvsall_mnist}
\end{figure}

\begin{figure}[t!]
    \centering
        \includegraphics[width=0.44\textwidth]{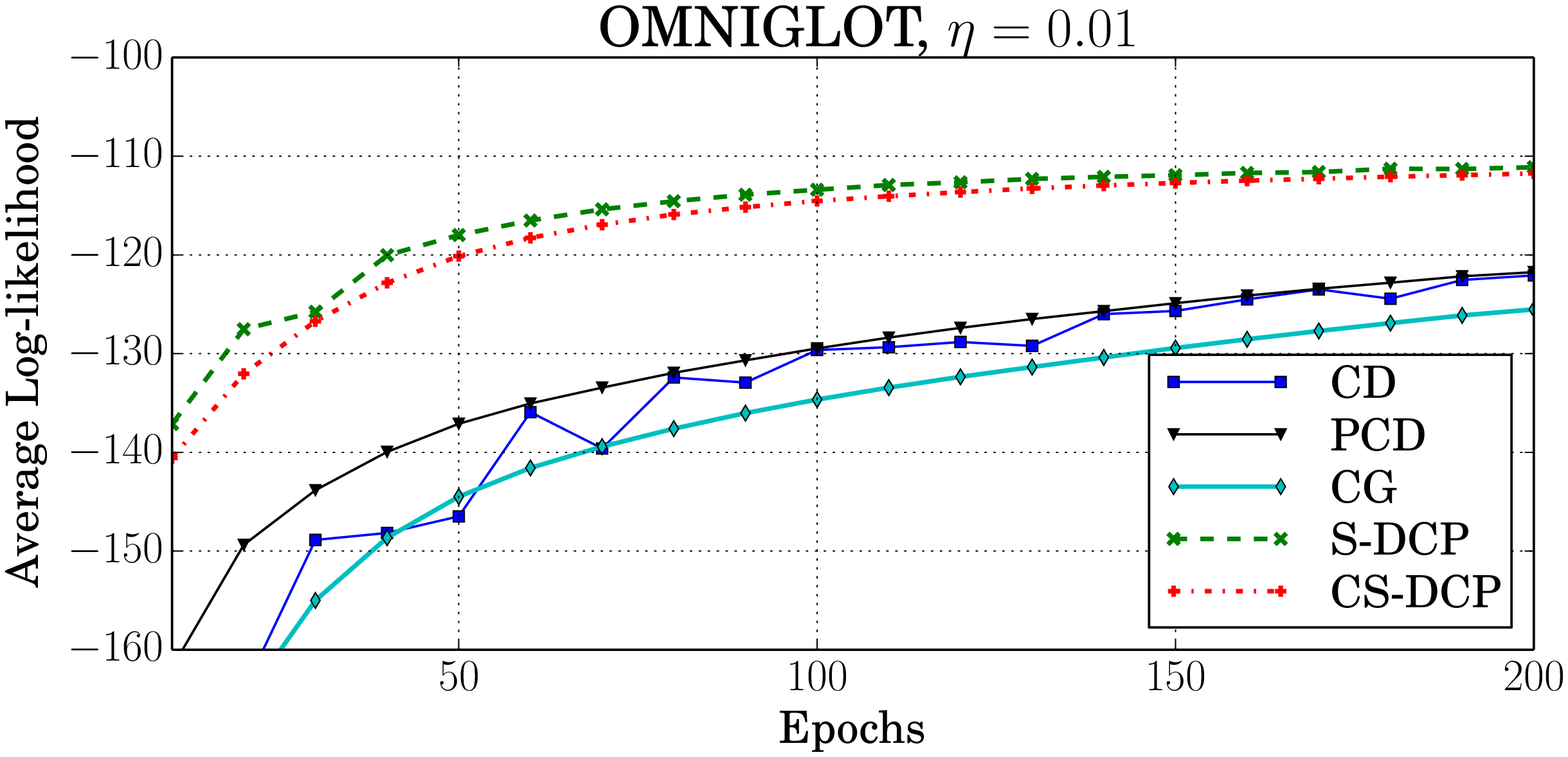}
        \includegraphics[width=0.44\textwidth]{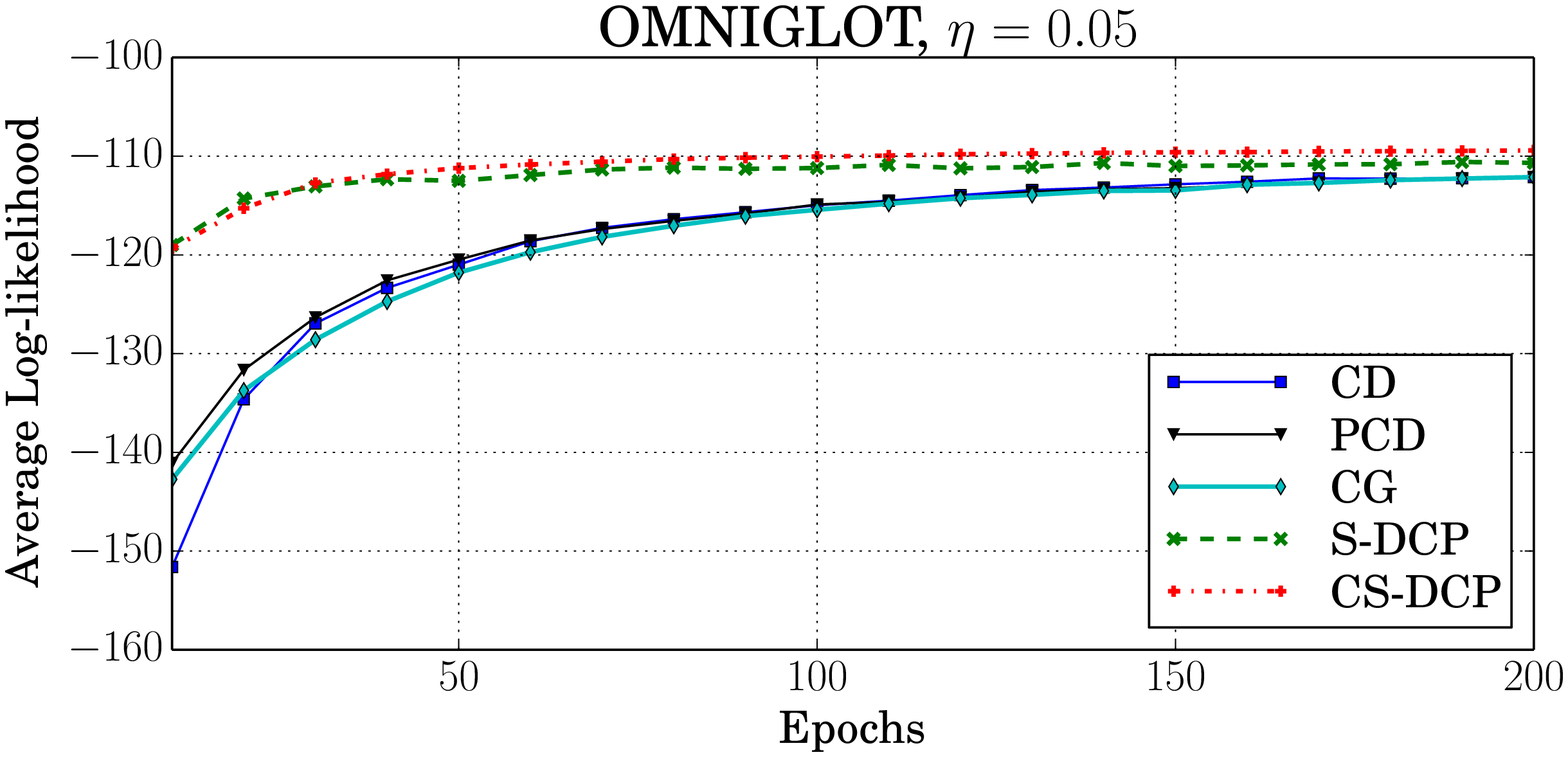}
    \caption{The performance of different algorithms on OMNIGLOT dataset}\label{SDCPvsall_omniglot}
\end{figure}

Fig. \ref{SDCPvsall_mnist} and Fig. \ref{SDCPvsall_omniglot}  show the results obtained using the MNIST and OMNIGLOT data sets respectively.
For the MNIST dataset, we observe that CS-DCP converges faster when a small learning rate is used. However, 
all the algorithms achieve almost the same ATLL at the end of $200$ epochs.
For the OMNIGLOT dataset, we see that both S-DCP and CS-DCP algorithms are superior to all the 
other algorithms both in terms of speed and the achieved ATLL.

The experimental results obtained on these two data sets indicate that
both S-DCP and CS-DCP perform well even when the dimension of the model is large. Their speed of learning
is also superior to that of other standard algorithms. For example, when $\eta=0.01$ the CS-DCP algorithm achieves $90\%$ of the maximum ATLL
approximately $100$ epochs before the $CG$ algorithm does as indicated by the vertical lines in Fig. \ref{SDCPvsall_mnist}.

\begin{figure}[t!]
    \centering
           \includegraphics[width=0.44\textwidth]{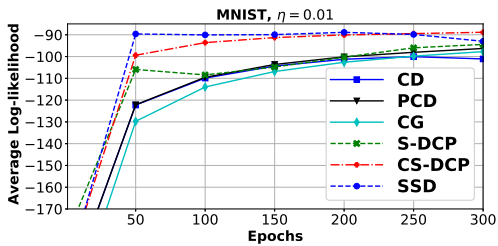} 
        \caption{Comparison of SSD with CS-DCP and S-DCP with a batch size of $1000$ on MNIST dataset.}\label{SDCP_SSD}
\end{figure}

As mentioned earlier, we use batch size $1000$ to compare the S-DCP and CS-DCP with the SSD algorithm.
We use exactly the same settings as used in  ~\cite{carlson2015stochastic} and report the performance on the    
statistically binarized MNIST dataset. For the S-DCP and CS-DCP, we use $d=6$ and $K'=4$ to match the CD steps of $K=25$ used in ~\cite{carlson2015stochastic}.
The results presented in Fig. \ref{SDCP_SSD} indicate that both CS-DCP and SSD have similar performance and are superior compared to all the other algorithms.
The SSD is slightly faster since it exploits the
specific structure in solving the optimization problem. However, as noted earlier SSD algorithm does not generalize well
for small data sets and also when small batch size is used for learning large data sets. Even for a large batch size
of $1000$ the divergence can 
be noted in Fig. \ref{SDCP_SSD} at the end of $250$ epochs. In addition, the SSD algorithm is computationally expensive.

\section{Sensitivity to hyperparameters}
\subsection{Effect of Learning Rate}
\begin{figure}
\centering
\begin{subfigure}
        \centering
 \includegraphics[width=0.44\textwidth]{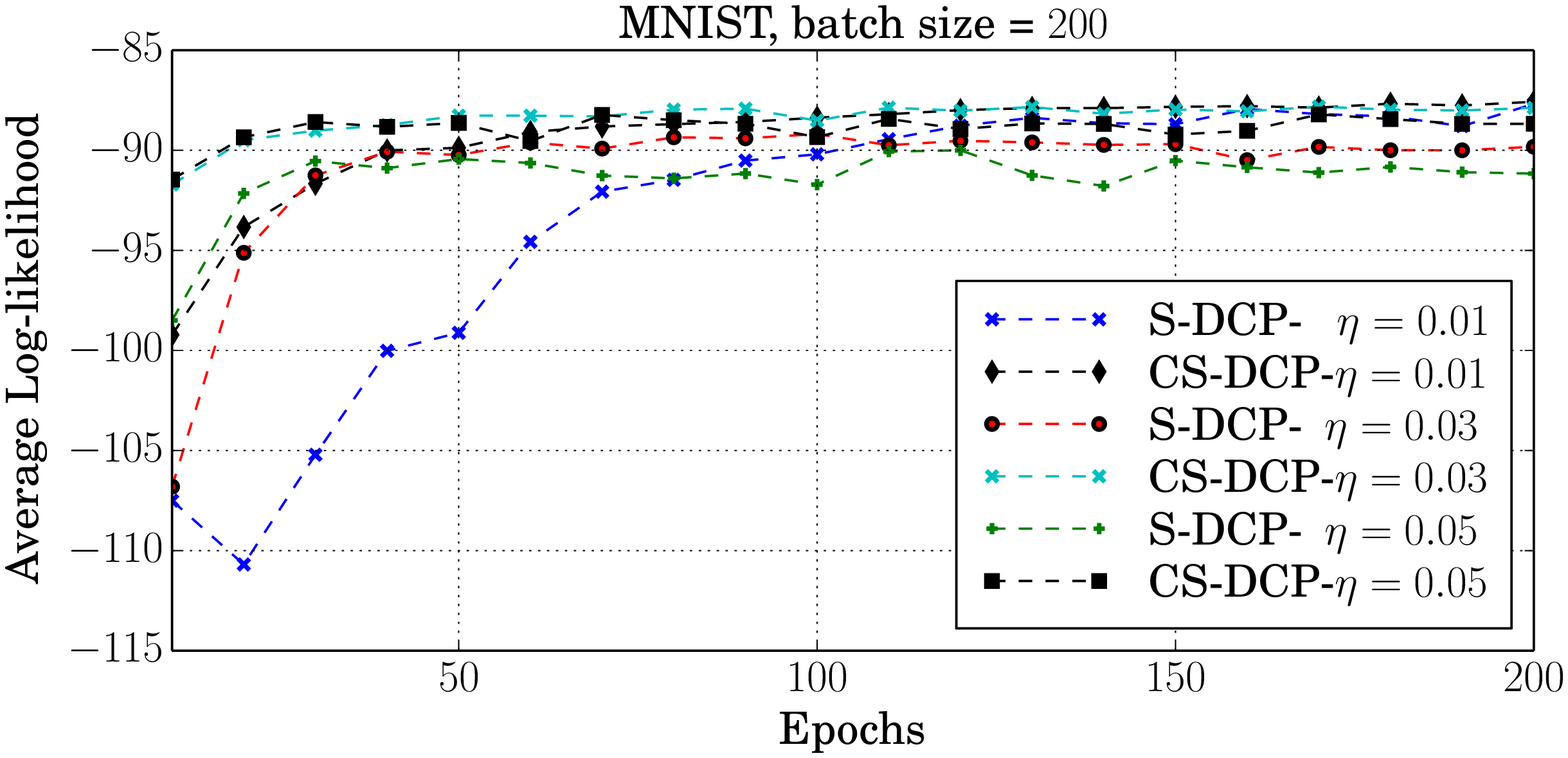}
 \end{subfigure}
\begin{subfigure}
        \centering
 \includegraphics[width=0.44\textwidth]{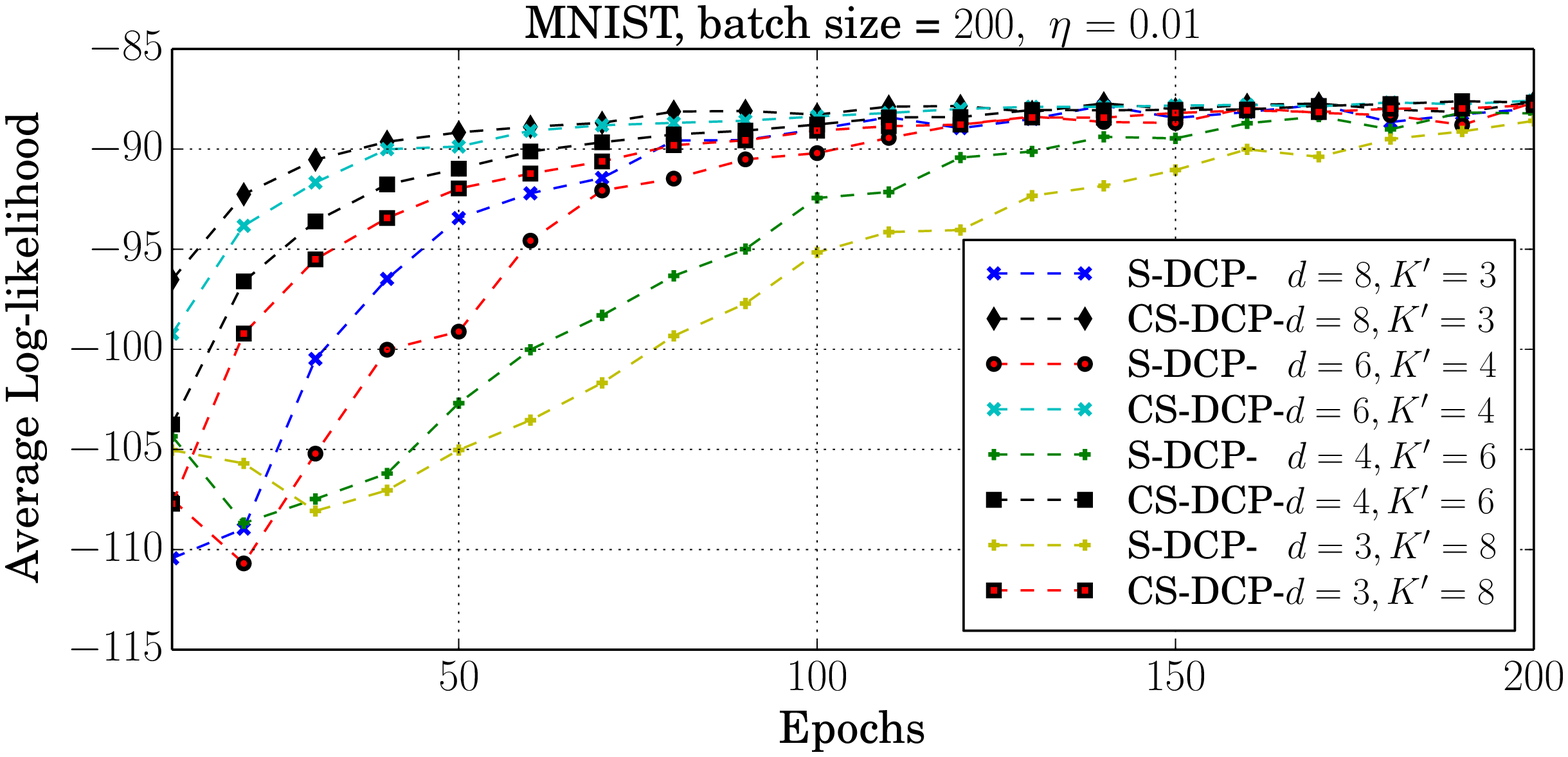}
 \caption{Comparison of S-DCP and CS-DCP algorithms with different (a) learning rates and (b) values of $d$ and $K'$ keeping $dK'=24$. }\label{mnist_var_dk}
  \end{subfigure} 
\end{figure}

The learning rate is a crucial hyperparameter in learning an RBM. If the learning rate is small we can ensure descent on each iteration. It can be seen from Fig. \ref{mnist_var_dk}(a)
that for a small learning rate $\eta=0.01$, as epoch progress, there is an increase in the ATLL for most epochs.
\subsection{Effect of $d$ and $K'$}
Fig. \ref{mnist_var_dk}(b) depicts the learning behavior of S-DCP and CS-DCP algorithms when the hyperparameters $d$ and $K'$ are varied, keeping the 
computational complexity per batch fixed.  
We fix $dK'=24$ and vary $d$ and $K'$. As $d$ is increased, the number of gradient updates per epoch increases leading
to faster learning indicated by the sharp increase in the ATLL. However, irrespective of the values $d$ and $K'$ 
both the algorithms converge eventually to the same ATLL. We find a similar trend in the performance across other data sets as well.

In a way, the above behavior is one of the major advantages of the proposed algorithm. For example, if a
longer Gibbs chain is required to obtain a good representative samples from a model, it is better to choose a smaller $d$ and a larger $K'$.
In such scenarios, the S-DCP and CS-DCP algorithms provide a better control on the dynamics of learning.

\subsection{Effect of Batch Size}
A small batch size is suggested in \cite{hinton2010practical} for better learning with a simple SGD approach.
However, in \cite{carlson2015stochastic}, advantages of using a large batch size of the order of $2m$ (twice the number of hidden units) with a 
well designed optimization, is demonstrated. If we fix the batch size, then the learning rate can be fixed through cross-validation.
However, to analyse the effect of batch size on S-DCP and CS-DCP learning, we fix the learning rate, $\eta=0.01$, and then vary the batch size.
As shown in Fig. \ref{mnist_var_init}(a), the S-DCP eventually reaches almost the same ATLL, after $200$ epochs of training, irrespective of the batch size considered.
Therefore, the proposed S-DCP algorithm is found to be less sensitive to batch size. We observed
a similar  behavior with the Omniglot dataset.

\subsection{Sensitivity to Initialization}
We initialize the weights with samples drawn from a Gaussian distribution having mean zero and standard deviation, $\sigma$. 
We consider a total of four cases: $\sigma=0.01$ and $\sigma=0.001$ and 
the biases of the visible units are initialized to zero or to the inverse sigmoid of the training sample mean (termed as base rate initialization). 
It is noted in \cite{hinton2010practical} that initializing the visible biases to the sample mean improves the learning behavior. 
The biases of the hidden units are set to zero in all the experiments. 

\begin{figure}
\centering
\begin{subfigure}
        \centering
 \includegraphics[width=0.44\textwidth]{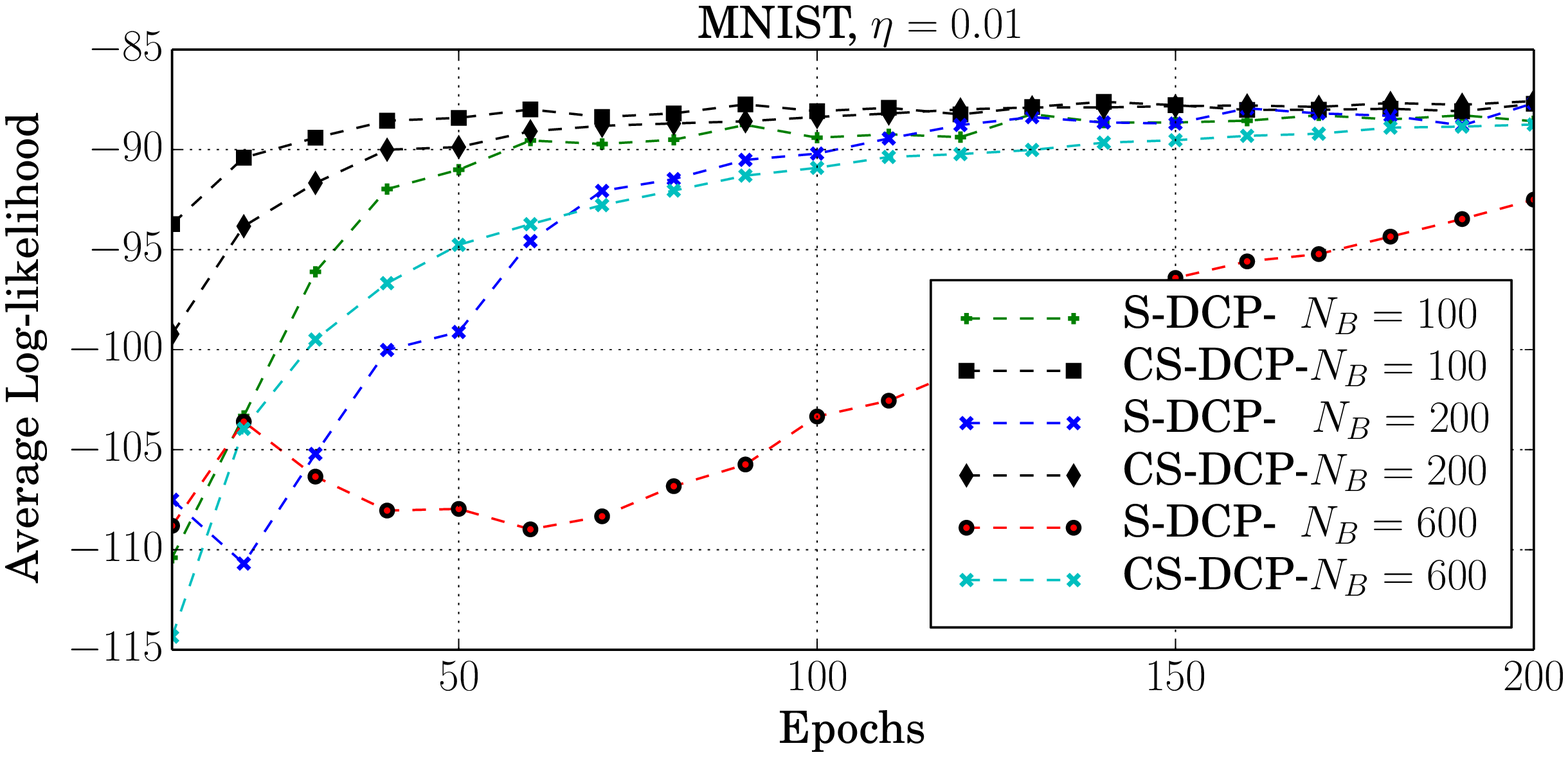}
 \end{subfigure}
 \begin{subfigure}
        \centering
 \includegraphics[width=0.44\textwidth]{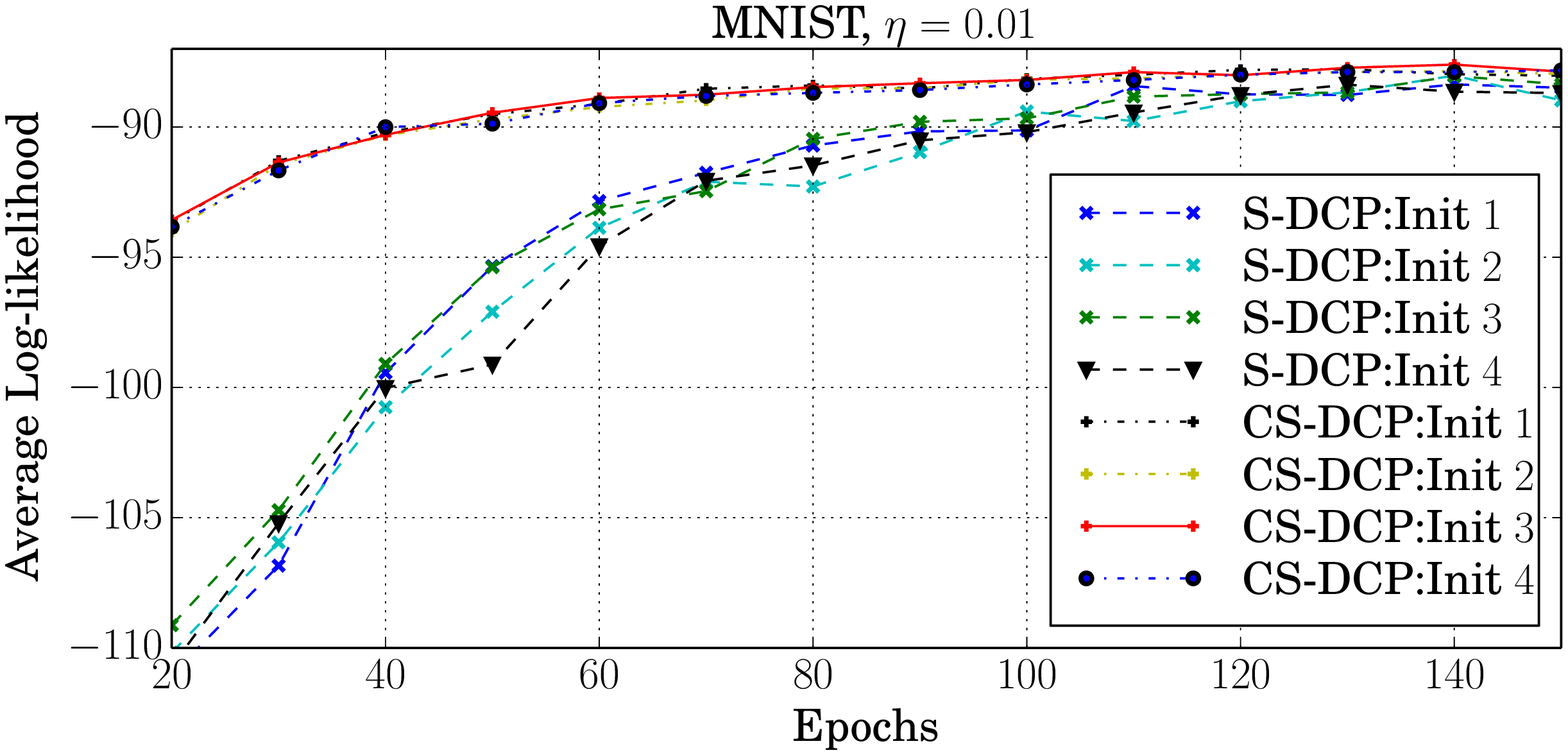}
 \end{subfigure}
\caption{MNIST dataset: Learning with different (a) batch sizes (b) Initialization}\label{mnist_var_init}
\end{figure}

The results shown in Fig. \ref{mnist_var_init}(b) indicate that both S-DCP and CS-DCP algorithms are less sensitive to the initial values of weights and biases.  
We observed similar behavior with the other datasets. 

 \begin{table}[ht!]                                                                                          
\centering                             
\caption{Legend Notation for Fig. \ref{mnist_var_init}(b).}\label{init_legend}
\small{
\begin{tabular}{|c|c|c|c|c|}
 \hline
 & Init $1$ & Init $2$& Init $3$& Init $4$\\
 \hline
 $\sigma$ & $0.001$ & $0.01$ & $0.001$ &$0.01$ \\
\hline
$\vecb$ & $\mathbf{0}$  & $\mathbf{0}$  & base rate & base rate\\
\hline
\end{tabular}
}
\end{table}
\section{Conclusions and Future Work}\label{sec:conclusions}
A major issue in learning an RBM is that the
 log-likelihood gradient is to be obtained with MCMC sampling and hence is noisy. There are several optimization techniques proposed to obtain fast and stable learning 
 in spite of this noisy gradient. 
In this paper, 
we proposed a new algorithm, S-DCP,
that is computationally simple but achieves higher efficiency of learning compared to CD and its variants, the current popular 
method for learning RBMs. 

We exploited the fact that the RBM log-likelihood function is a difference of convex functions and adopted
the standard DC programming approach for maximizing the log-likelihood. We used SGD for solving the convex optimization
problem, approximately, in each step of the DC programming. The resulting algorithm is very similar to CD and has the same computational complexity. As a matter of fact,
CD can be obtained as a special case of our proposed S-DCP approach. Through extensive empirical studies, we illustrated the advantages of S-DCP over CD.
We also presented a centered gradient version CS-DCP. We showed that S-DCP/CS-DCP is very resilient to the choice of hyperparameters through simulation results. 
The main attraction of 
S-DCP/CS-DCP, in our opinion, is its simplicity compared to other sophisticated optimization techniques that are proposed in literature for this problem.

The fact that the log-likelihood function of RBM is a difference of convex functions, opens up new areas to explore, further.
For example, the inner loop of S-DCP solves a convex optimization problem using SGD. A proper 
choice of an adaptable step size could make it more efficient
and robust. Also, there exist many other efficient techniques for solving such convex optimization problems.
Investigating the suitability of such algorithms for the case where we have access only to noisy gradient values (obtained through MCMC samples)
is also an interesting problem for future work.

\acks{We thank the NVIDIA Corporation for the donation of the Titan X Pascal GPU used for this research.}

\def\url#1{}
\def\doi#1{}

\bibliography{conference}
\end{document}